\documentclass{article}

\usepackage{times}
\usepackage{soul}
\usepackage{url}
\usepackage{placeins}
\usepackage[hidelinks]{hyperref}
\usepackage[utf8]{inputenc}
\usepackage[small]{caption}
\usepackage{graphicx}
\usepackage{amsmath}
\usepackage{amsthm}
\usepackage{booktabs}
\usepackage{algorithm}
\usepackage{algorithmic}
\usepackage{multirow}
\usepackage{todonotes}
\newcommand{\figSH}{1.0in}

\title{Injective Domain Knowledge in Neural Networks for Transprecision
Computing}
\date{}
\author{
    Andrea Borghesi, Federico Baldo, Michele Lombardi, Michela Milano
    \\ DISI, University of Bologna
}

\begin{document}

\maketitle

\begin{abstract}
    Machine Learning (ML) models are very effective in many learning tasks, due
to the 
    capability to extract meaningful information from large data sets.
Nevertheless, 
    there are learning problems that cannot be easily solved relying on pure
data,
    e.g. scarce data or very complex functions to be approximated.
    Fortunately, in many contexts domain knowledge is explicitly available 
    and can be used to train better ML models. 
    
    This paper studies the improvements that can be obtained by integrating
prior
    knowledge when dealing with a non-trivial learning task, namely precision
tuning of 
    transprecision computing 
    applications. The domain information is injected in the ML models in
different ways:
    I) additional features, II) ad-hoc graph-based network topology, III) 
    regularization schemes.
    The results clearly show that ML models exploiting problem-specific
information 
    outperform the purely data-driven ones, with an average accuracy improvement
around 38\%.
\end{abstract}

\section{Introduction}
In recent years, ML approaches have been exhaustively proved to be successful 
with a wide range of learning tasks. Typically, ML models are sub-symbolic,
black-box techniques 
capable of effectively exploiting the information contained in large amount of
data. 
Part of their usefulness is their adaptability, that is the fact that ML models 
with the same architecture and training algorithm 
can be applied in very different contexts with good results. This happens
because
most ML approaches make very few assumptions on the underlying data and the
functions 
that they are trying to learn.

However, 
purely data-driven models can be not ideal if, for instance, the data is
relatively 
expensive to obtain and the function to be learned is very hard.
At the same time, in many areas domain-specific information is available (e.g.
structured
data, knowledge about the data generation process, domain experts experience,
etc)
but not exploited. 
In such cases, it makes sense to take advantage of this information to improve
the performance
of the ML techniques, so they do not have to start from scratch while dealing
with difficult learning tasks. In other words, \emph{why learn again something
that 
you already know?}.

In this paper we discuss a strategy to inject domain knowledge expressed as
constraints
in a ML 
model, namely a Neural Network (NN). We consider different sources of prior
information 
and adopt suitable injection approaches for each of them: I) feature extraction, 
II) ad-hoc NN structure, and III) data augmentation combined with a
regularization 
strategy.
As a case study, we selected a complex supervised learning problem drawn from
the area of 
\emph{transprecision computing}, a novel paradigm that allows trade-offs between 
the energy required to perform the computation and the accuracy of its 
outcome~\cite{malossi2018transprecision}. The learning task is very 
hard, due to non-linearity, non-monotonicity and relatively small data sets (a
few 
thousands of samples). 
The experimental 
results clearly show that exploiting prior information leads to remarkable
gains.
On average over all benchmarks, the knowledge injection provides a 
 38\% improvement in terms of prediction accuracy.
The rest of the paper is structured as follows: after the discussion about
related 
works (Section~\ref{sec:related}), Section~\ref{sec:approach} introduces the
injection 
approaches; Sec.~\ref{sec:trans} details
 transprecision computing and the specific learning task, highlighting its
difficulty and the domain knowledge that can be extracted;
Sec.~\ref{sec:exp_eval} summarizes the 
experimental results.; finally, Sec.~\ref{sec:conclusion} concludes the paper.

\section{Related Works}
\label{sec:related}

The combination of sub-symbolic models with domain knowledge is an area explored
by previous 
research~\cite{song2017machine}
in many fields~\cite{yu2010vqsvm,childs2019embedding,miao2019leveraging},
ranging from strategy games~\cite{nechepurenko2019comparing} to fall detection
systems~\cite{mirchevska2014combining}, 
and several different techniques have been proposed. For instance, feature
engineering~\cite{zheng2018feature} is a common method for improving the
accuracy of 
purely data-drive ML models by selecting useful features and/or transform the
original 
ones to facilitate the learner's task. In general, this is not a trivial problem
and 
requires much effort, both from system expert and ML
practitioners~\cite{khurana2018feature}.
In this paper, we employ a slightly different approach, as we employ domain
knowledge
to create novel features that render \emph{explicit} the information hidden in
the 
raw data.

Another research direction aims at training NNs while forcing constraints which
can be 
drawn from knowledge domain. \cite{fischer2018dl2} present
a method for translating logical constraints in loss functions that guide the
training 
towards the desired output. \cite{muralidhar2018incorporating} propose a
different approach
to incorporate domain knowledge in a NN by adopting a loss function that merges
mean squared error and a penalty measuring whether the NN output respect a set
of constraints
derived from the domain; the method is limited to constraints enforcing
monotonicity and bounds 
on the target variable. \cite{xu2017semantic} introduce a method to integrate
semantic 
knowledge in deep NNs, again exploiting a loss function; in this case the
approach is 
targeted at semi-supervised learning and not well suited for supervised tasks.
Acting on the loss function with a regularization term has been proposed also by
\cite{diligenti2017semantic}, with their work on Semantic-Based Regularization
(SBR), 
a method to merge high-level domain information expressed as first-order logic
in ML models.
We have exploited their technique in combination to a data augmentation strategy
to enhance a ML model.
\cite{li2019augmenting} propose to integrate first-order logic in NNs by adding 
non-trainable layers and neurons that represent the logic predicates; their
approach focuses
explicitly on NLP tasks.

Graph Convolutional Neural Networks (GCNN)
\cite{Kipf:2016tc,defferrard2016convolutional} are a type of neural 
networks specialized for learning tasks involving graphs. GCNNs have
been recently used in several fields~\cite{zhang2019graph}, owning to their
capability
to deal with data whose structure can be described via graphs, thanks to a
generalization 
in the spectral domain of the convolutional layers found in many deep learning
networks. 
GCNNs most common applications involve semi-supervised classification tasks,
with the goal
of predicting the class of unlabeled nodes in a graph -- a case of graph
learning. 

\section{Domain Knowledge Injection}
\label{sec:approach}
The main goal of this paper is the exploration of how a ML model can be improved 
through the exploitation of domain knowledge.
We claim that purely data-driven ML models can benefit from the injection of
prior 
knowledge provided by domain experts; Sec.~\ref{sec:exp_eval} will report 
the results of the experimental evaluation, conducted on a specific learning
task
where domain knowledge is available (details in Sec.~\ref{sec:trans}).
We consider domain knowledge that can be expressed in the form of logical 
constraints between variables (input and output features of the ML models)
and/or encoded 
in a graph.

Let $X$ be the training set and $y$ the targets, either continuous values
(regression) 
or categorical labels (classification), $f$ the model trained to learn the
relation between $X$ and $y$.
In general, domain knowledge can be expressed as a set of logical constraints
between 
the input features $X$ and the target $y$. For instance, the monotonicity
propriety holds 
if $x_1 \leq x_2 \implies y_1 \leq y_2$ for every pair in the $X$. 
We propose a multi-faceted domain knowledge injection strategy and we 
 introduce three different approaches, 
each one addressing a specific weakness encountered by purely data-driven
techniques:
\begin{enumerate}
\item \emph{feature extraction} for information implicit but hidden in the raw
data
-- if the examples available in the data set are not sufficient nor informative
enough to train accurate ML models, a set of additional features can be 
created using the domain knowledge and reasoning about the relationships among 
the original features;
\item \emph{ad-hoc network topology} for learning tasks where the relationships 
among the features and the data structure can be encoded with graphs;
\item \emph{data augmentation and regularization function} for a twofold scope:
I)
learning with very few data (e.g. active learning), II) enforcing desired
proprieties 
in the output of the ML model.
\end{enumerate}
The feature extraction (1) takes into account prior knowledge that can be
expressed via
a set of binary constraints $C$ among the input features $X$; these constraints
can be 
used to obtain an extended training set $X'$ by checking if every example in $X$
satisfies
them or not.
The regularization method (3) assumes that the knowledge can be expressed as
first-order
logic constraints between input features $X$ and the target $y$; data
augmentation helps to cope with scarce data and amplify the effect of the
regularization.

We introduce the knowledge injection strategy and present three 
different techniques, each tailored for a specific source of information. At the 
current stage we were more interested in measuring the specific contribution of 
each method, thus they were tested separately, but we plan to explore hybrid 
solutions in future works.
As a case study we consider a complex supervised learning task and then we
tackle 
it with multiple purely data-driven ML models, and in particular we use neural
networks (NN).
Subsequently, we inject the domain knowledge and then we experimentally evaluate
the 
obtained improvements.

\section{Transprecision Computing}
\label{sec:trans}
There exist many techniques for transprecision computing and in this paper we
focus on an approach targeting floating-point (FP) 
variables and operations, as their execution and data transfer can 
require a large share of the energy consumption for many applications; 
decreasing the number of bits used to represent FP variables can lead to 
energy savings, with the side-effect of reduced accuracy on the outcome of 
the application (also referred to as \emph{benchmark}).
Deciding the optimal number of bits for 
FP variables while respecting a bound on the computation accuracy is referred 
to as \emph{precision tuning}. In this context, understanding the relationship 
between assigned precision and accuracy is a critical issue, and not an easy
one, as this relationship cannot be analytically expressed for non-trivial 
benchmarks~\cite{moscato2017automatic}. Therefore, we address this problem 
via a ML model, that is \emph{learning} the relationship between precision 
and accuracy. For this scope, we use a transprecision library for 
precision tuning called \emph{FlexFloat}~\cite{tagliavini2018transprecision} 
to create a suitable data set; this means running a benchmark with multiple
precision configurations and store the associated error. 
As this is a highly time consuming task, we work with data sets of relatively 
limited size (5000 samples at maximum)
\footnote{The learning task is only a part of a larger project aiming at solving
the precision tuning problem with optimization techniques; state-of-the-art
algorithms for FP precision tuning (e.g. \cite{ho2017efficient}) dictate a bound
on the time to solve the optimization problem -- hence, the need of a low
data set creation time}
, an issue that complicates the learning task.

\subsection{Problem Description}
\label{sec:trans_problem_description}
We consider numerical benchmarks where multiple FP variables partake in the 
computation of the result for a given input set, which includes a structured 
set of FP values (typically a vector or a matrix). The number of variables 
with controllable precision in a benchmark $B$ is $n_{var}^B$; these variables 
are the union of
the original variables of the program and the additional variables inserted 
by FlexFloat to handle intermediate results (see
\cite{tagliavini2018transprecision}
for details).
FlexFloat allows to run a benchmark with different precision (different numbers
of bits assigned to the FP variables) and to measure the reduction in output
quality
due to the adjusted precision
(reduction w.r.t. the output obtained with maximum precision) -- we will refer
to
this reduction as \emph{error}. If $O$ indicates the result computed with the
tuned 
precision and $O^M$ the one obtained with maximum precision, the error $E$ is
given by 
$E = \max_{i} \frac{(o_i - o_i^M)^2}{(o_i^M)^2}$ -- this is the metric adopted 
in the transprecision
community~\cite{ho2017efficient,malossi2018transprecision}.
As a case study we selected a representative subset of the benchmarks studied in 
the context of transprecision computing~\cite{oprecomp_project}. At this stage,
we do 
not focus on whole applications but rather on micro-benchmarks, in particular
the 
following ones:
    1) \emph{FWT}, Fast Walsh Transform for real vectors, from
    the domain of advanced linear algebra ($n_{var}^{FWT} = 2$); 
    2) \emph{saxpy}, a generalized vector addition
    , basic linear algebra ($n_{var}^{saxpy} = 3$); 
    3) \emph{convolution}, convolution of a matrix, ML ($n_{var}^{conv} = 4$); 
    4) \emph{dwt}, Discrete wavelet
    transform, from signal processing ($n_{var}^{dwt} = 7$); 
    5) \emph{correlation},
    compute correlation matrix of input, data mining ($n_{var}^{corr} = 7$).
    6) \emph{BlackScholes}, estimates the price for a set of options applying
    Black-Scholes equation, from computational finance ($n_{var}^{BScholes} =
    15$).

Beside the precision configuration, another element that impacts a benchmark's 
output, and thus the error, is the input set fed to the application (e.g, the
actual 
values of the FP variables). The vast majority of transprecision tuning
approaches 
consider the single input set case~\cite{ho2017efficient}:
a fixed input set is given to the benchmark and the precision of the variables
is tuned for that particular input set (no guarantee that the configuration 
found will suit different input sets).
We opted for ``stochastic'' approach: 
we consider multiple input sets, so that a distribution of errors
is associated to each configuration, rather than a single value.
The learning task is then not to predict
the error associated to a specific input set but to learn the relation
between precision configuration and \emph{mean error} over all input sets.
Learning the relationship between variable precision
and error is a hard problem. First, the error metric is very susceptible to 
differences between output at maximum precision and output at reduced precision, 
due to the maximization component. Secondly, the precision-error space is 
non-smooth, non-linear, non-monotonic, and with many
peaks (local optima). In practice, increasing the precision of all variables
does not guarantee an error reduction. This effect is due to multiple factors,
from
the impact of rounding operations to the effects of numerical stability on the
control flow~\cite{Darulova2017}.  

\subsection{Data Set Creation}
\label{sec:trans_dataset}
As a first step, we created a collection of data sets containing examples of the
benchmarks run at different precision, with the corresponding error values. 
We call \emph{configuration} the assignment of a precision
to each FP variable.
The configuration space was explored via Latin Hypercube
Sampling (LHS)~\cite{stein1987large}. 
As described in the previous section, for each configuration the benchmarks 
were run with 30 different input sets\footnote{Long vectors 
and matrices containing different real values} and the error associated to each
combination of \textless configuration, input set\textgreater~ was computed.
As target we then use the average over the 30 input-specific errors.
The
majority of configurations lead to small errors, from $10^{-1}$ to $10^{-30}$.
However, in a minority of cases lowering the precision of critical variables
generates extremely large errors; in the transprecision computing context,
error larger than 1 are deemed excessive.

After a preliminary analysis, we realized that for a ML model is very hard to 
discern between small and relatively close errors (i.e. $e-^{20}$ and
$e-^{15}$);
we therefore opted to predict the negative of the logarithm of the error, thus
magnifying
the relative differences. 
Moreover, a careful examination revealed that overly large $E$ values were
usually due to
numerical issues arising during computation (e.g. overflow, underflow, division
by zero, or not-a-number exceptions). This intuitively means that the
large-error configurations are likely to follow a distinct pattern w.r.t. the
configurations having a more reasonable error value. We are much more interested
in 
relatively small error (e.g. $E \leq 0.95$, not in logarithmic scale) as in 
transprecision computing the largest accepted error is typically $0.1$ 
(meaning an output accuracy higher than 90\%). 
Hence, we decide to level out all the errors in the data set above the $0.95$
threshold; 
if the \textless configuration, input set\textgreater~combination produced an
error
 $E \geq 0.95$, after pre-processing its error is set to $0.95$ (before the
conversion to
 logarithmic scale).

\subsection{Knowledge Injection}
\label{sec:trans_domain_prior}

\begin{figure}[hbt]
    \centering
    \includegraphics[width=0.25\textwidth,height=\figSH]{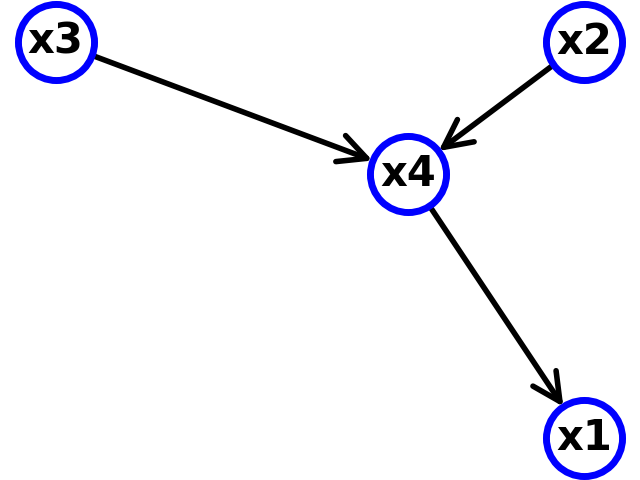}
    \caption{Example of Dependency Graph}
    \label{fig:depGraph_example}
\end{figure}

As the benchmarks are programs composed by a set of 
interdependent FP variables, the variables' interactions represent a source
of valuable information for learning the relationship between precision and
error.
This domain-level knowledge is encoded in the \emph{dependency graph} of the
benchmark, which specifies how the program variables are related. 
For instance,
consider the expression $V_1 = V_2 + V_3$; this corresponds to four
precision that need to be decided $x_i, i \in [1,4]$. The first three
precision-variables $x_1$, $x_2$, and $x_3$ represent the precision of the
actual variables of the expression, respectively $V_1$, $V_2$, and $V_3$; the
last variable $x_4$ is a \emph{temporary} variable introduced by
FlexFloat to handle the (possibly) mismatching precision of the
operands $V_2$ and $V_3$ (FlexFloat performs a cast from $x_2$ and $x_3$
to the intermediate precision $x_4$). Each variable is a node in the dependency
graph, and the relations among variables are directed edges, as depicted in
Fig.~\ref{fig:depGraph_example}; an edge entering a node means that the
precision of the source-variable is linked to the precision of the
destination-variable.

\subsubsection{Additional Features Extraction}
\label{sec:trans_additional_features}
As we have seen, the prior 
information on the benchmarks is encoded in directed graphs; for explanatory
purposes, 
we will take as example the micro-benchmark represented by the graph in 
Fig.~\ref{fig:depGraph_example}.
Using the encoded knowledge, a set of additional features characterizing 
the precision configurations can be obtained.
We consider only one type of relation, that is 
assignments (e.g. $x_4 \rightarrow x_1$). In this kind of expression, granting a 
larger number of bits to the value to be
assigned $x_4$ would be pointless since the final precision of the expression is
ultimately governed by the precision of the result variable $x_1$.  
Configurations that respect this relationship have a higher probability 
to lead to smaller errors w.r.t. configurations that do not respect this 
 constraint. In practice, configurations where $x_4 \leq x_1$ are 
associated to smaller errors.  

This information can be added to the training set as a 
collection of additional features. 
For each couple of variables involved in an 
assignment operation $x_i \rightarrow x_j$ we compute the feature $F_{ji} = x_j
- x_i$ 
\footnote{If $F_{ji} \leq 0$ it means that the $x_i \leq x_j$ is not respected,
hence 
a higher error is associated to the configuration}, which is then added to the
data set. 
Each feature corresponds to one of the logic binary constraints used to express
the domain knowledge.
For instance, if we consider again the example of
Fig.~\ref{fig:depGraph_example} 
there are three additional features, one for each assignment expression
(highlighted
by the three arrows in the graph): $F_{43} = x_4 - x_3$, $F_{42} = x_4 - x_2$,
$F_{14} = x_1 - x_4$. 
Thanks to these additional features 
 an extended data set can be obtained. If we consider two
 possible configurations for the micro-benchmark in
Fig.~\ref{fig:depGraph_example},
$C_1 = [27, 45, 35, 40]$ and $C_2 = [42, 23, 4, 10]$, the original data set
would be
composed by four features (one for each FP variable) plus the associate error
(the 
target of the regression task). Instead, the extended data set contains seven
features 
plus the error: $C_1^{ext} = [27, 45, 35, 40, 13, -5, -5]$ and
$C_2^{ext} = [42, 23, 4, 10, -32, -13, -6]$.

\subsubsection{Graphical Convolutional Neural Networks}
\label{sec:trans_gcnn}

The transprecision learning task is a supervised regression problem whose prior
information can be expressed through a graph, that is the dependency graph that
links
the variable in the benchmark. As mentioned in Section~\ref{sec:approach}, GCNNs
are
well suited to deal with graph-structured problems.
As our problem is slightly different from those considered in the literature, 
we did not adopted the standard approach but we rather exploited the main
component of GCNNs, the 
\emph{graph convolution}, implemented via Graph Convolutional Layers (GCL), and
applied to
the transprecision task.
The GCNN has the following structure:
first, from the dependency graph we compute the adjacency matrix; then the
adjacency matrix
and the input feature matrix are combined to form the input of a first
GCL, which is then fed to a second one. Its output becomes the input for a fully 
connected dense layer with 128 neurons, followed by two other fully connected
layers of
decreasing dimension (respectively, 32 and 8).
The final layer is, again, a dense layer with a single neuron, that is the
network output.

\subsubsection{Data Augmentation and Regularization}
\label{sec:trans_SBR} 

As mentioned in Sec.~\ref{sec:trans}, the learning task is made more difficult
by 
the presence of non-mononicity: situations where the normal precision-error
relationship is 
not respected. 
They arise due to numerical instability, and their presence
is magnified by the use of small data sets and a limited number of different
input sets; with
sufficiently large data sets they would be discarded as outliers. 
As mentioned before, the learning task addressed in this paper is a step towards
an
optimization model for precision tuning; with this scope in mind, it would be 
preferable to have a ML model that does not reproduce non-monotonicity events in
its 
predictions.
This is a domain knowledge about an undesirable propriety that should be
corrected. 
The problem with non-monotonicity would be solved if we could have more training
examples, 
but this is not easily attainable as we should run a benchmark to compute the
error 
associated to a configuration. 
However, generating new configurations without computing the error is trivial;
we can
exploit this advantage in conjunction with an appropriate regularization scheme
in 
order to impose monotonicity on the ML model predictions. This process is a
form of \emph{data augmentation}.
Injecting the monotonicity constraint in the training process may allow to
mitigate the noise and improve generalization, even with smaller training sets.
We take into account such constraint at training time by exploiting ideas from
Semantic Based Regularization \cite{diligenti2017semantic}, an approach that
advocates to the use of (differentiable) constraints as regularizers in the loss
function. Let us write $x_i \prec x_j$ if configuration $x_j$ dominates $x_i$,
i.e. if every variable in $x_j$ has precision higher than or equal to $x_i$; let
$P$ be the set of dominated-dominating pairs in our training set $X$, $P =
{(i,j) |x_i \prec x_j}$. Then, we can formulate the following regularized loss
function:
\begin{equation}
    MSE(X, y) + \lambda \sum_{i, j \in P} max(0, f(x_j) - f(x_i))
\end{equation}
where $f$ is the error predictor being trained, and MSE is the mean squared
error. Each regularization term is associated to a pair in $P$ and has non-zero
value iff the error for the dominating pair is larger than for the dominated
pair, i.e. if the monotonicity constraint is violated. New configurations in $P$
can be generated in order to get a much stable regularization factor without the
need of a bigger train set

Is worth noticing that SBR is orthogonal to the use of additional features 
hence the two methods can be combined; we plan to explore the benefits of
merging multiple 
methods in future works.

\section{Experimental Evaluation}
\label{sec:exp_eval}
We selected 5 different purely data-driven models to obtain a baseline:
I) a black-box optimization method (\emph{AutoSklearn}); 
II) a NN composed of 4 dense layers with $10 \times n_{var}^B$ neurons each,
that 
    is, the number of variables in a benchmark multiplied by 10 (\emph{NN-1});
III) a NN composed of 4 dense layers with $100 \times n_{var}^B$ neurons each
(\emph{NN-2});
IV) a NN composed of 10 dense layers with $10 \times n_{var}^B$ neurons each
(\emph{NN-3});
V) a NN composed of 20 dense layers with $10 \times n_{var}^B$ neurons each
(\emph{NN-4}).
All NNs have a single-neuron output layer fully connected 
with the previous one.
The black-box method used was drawn from the \emph{AutoML} area, 
namely a framework called \emph{autosklearn}~\cite{autosklearn} which uses 
Bayesian optimization for algorithm configuration and selection (e.g. finding
the best set of hyperparameters for a given task).
Our problem can be cast in the AutoML mold if we treat 
the variables precision as the algorithm configurations to be explored and 
the associated computation error as the target.

The code used to run the experiments was written in Python, using Keras and
TensorFlow 
for the implementation of the neural networks. \emph{Autosklearn} is distributed
as a Python
library and we used the version available online
\footnote{https://automl.github.io/auto-sklearn/master/}, with default
parameters and letting the
framework choose among all the implemented regression models.
The GCNN model was created using the \emph{Spektral} library
\footnote{https://danielegrattarola.github.io/spektral/}.
All the results presented in this section were run on 20 different instances
(different 
training and test sets) and we report the average values. Both input feature and 
targets were normalized. The code used to run the experiments is available at
this 
online repository [REDACTED\_FOR\_BLIND\_REVIEW].


To evaluate the impact of the additional features, the four different neural
networks
previously defined (NN-1, NN-2, NN-3, NN-4) were trained and tested both with
and 
without the extended data set.
At this stage we focus on the number of layers and their width and discarded
other
hyperparameters; their exploration will be the subject of future research works.
In this paper, these are the values for the main hyperparameter used with all 
methods: number of epochs $ = 1000$; batch size $ = 32$; as training algorithm 
we opted for \emph{Adam}~\cite{kingma2014adam} with standard parameters;
Mean Squared Error as loss function.
The data augmentation and SBR approach is used on top of a neural network with
the
same number of layers and neurons as NN-1.
The new configurations are injected in each batch during the training, with a
fixed size of 
$256$ elements; the amount of data generated is specified by a ratio, which
represents the 
percentage of samples introduced by the data augmentation.

\subsection{Models Accuracy}
\label{sec:exp_eval_MAE}

We begin by evaluating the prediction accuracy of the proposed approaches. We
measure the
accuracy using the Mean Average Error (MAE).
In Table\ref{tab:mae_comparison} we compare the results obtained using a
training and test 
set size of, respectively, 5000 and 1000 examples; test and training set 
are randomly drawn from the samples generated through LHS.
The first column of the table identifies the benchmark (the last row corresponds
to 
the average over all of them); the second column contains the MAE obtained with
the black-box 
approach, \emph{AutoSklearn}; columns 3 and 4 report the MAE with the first NN
(NN-1), respectively
without and with the additional features; the three following couples of columns
are the results with the other NNs (NN-2, NN-2, NN-3), again split between base
and 
extended data set; the final two columns correspond respectively to MAE obtained
with GCNN and  with SBR.
For this table, we consider the SBR approach with 75\% of augmented examples --
more details
 at Sec.~\ref{sec:exp_eval_SBR}.
        
\begin{table*}
\scriptsize\sf\centering
\begin{tabular}{l cc cc cc cc ccc}
\toprule
\multirow{2}{*}{\emph{Benchmark}} & \multirow{2}{*}{AutoSklearn} &
\multicolumn{2}{c}{NN-1}  & \multicolumn{2}{c}{NN-2} 
& \multicolumn{2}{c}{NN-3} & \multicolumn{2}{c}{NN-4} & \multirow{2}{*}{GCNN} & 
  \multirow{2}{*}{SBR} \\
 & & Base & Ext. & Base & Ext. & Base & Ext. & Base & Ext. & & \\
\midrule
\emph{FWT}          & 0.394 & 0.315 & 0.251 & 0.056 & 0.054 & 0.104 & 0.061 &
0.070 & 0.105 & 0.351 & 0.243 \\ 
\emph{saxpy}        & 0.003 & 0.000 & 0.000 & 0.000 & 0.000 & 0.000 & 0.000 &
0.000 & 0.000 & 0.000 & 0.001 \\
\emph{convolution}  & 0.020 & 0.005 & 0.005 & 0.002 & 0.002 & 0.003 & 0.003 &
0.003 & 0.003 & 0.004 & 0.006 \\
\emph{correlation}  & 0.397 & 0.139 & 0.120 & 0.091 & 0.092 & 0.111 & 0.098 &
0.114 & 0.102 & 0.262 & 0.139 \\
\emph{dwt}          & 0.422 & 0.057 & 0.034 & 0.011 & 0.012 & 0.029 & 0.020 &
0.031 & 0.022 & 0.072 & 0.068 \\
\emph{BlackScholes} & 0.411 & 0.238 & 0.047 & 0.184 & 0.035 & 0.239 & 0.038 &
0.297 & 0.172 & 0.307 & 0.220 \\
 \midrule
\emph{Average}      & 0.274 & 0.126 & 0.076 & 0.057 & 0.033 & 0.081 & 0.037 &
0.086 & 0.067 & 0.166 & 0.113 \\  
\bottomrule
\end{tabular}
\caption{Knowledge injection approaches comparison: Mean Average Error -- train
set size: 5k}
\label{tab:mae_comparison}  
\end{table*} 

The black-box model \emph{AutoSklearn} has clearly the worst performance, which
is not 
entirely surprising given the complexity of the learning task. The first
unexpected
and disappointing result is the poor performance of the GCNN, that is
outperformed 
by all other approaches in almost all benchmarks. We remark that this was a
novel 
application of GCNN and this preliminary analysis merely suggests that a more 
careful exploration is needed. Changing the network type can produce good
results: using a wider NN
(from NN-1 to NN-2) greatly reduces the MAE, while deeper NNs
provide smaller improvements (e.g. NN-3 and NN-4).
Very interestingly, a major MAE reduction is obtained by using the 
additional features (column \emph{Ext.}): for all NN types and over all
benchmarks,
the approach using the extended data greatly outperforms the baseline, with an 
average improvement of 39.7\% (considering all four NN types). 
The results obtained with data agumentation and SBR show that this method
performs better than \emph{AutoSklearn}
and the simplest NN without the additional features (NN-1), but it has a higher 
MAE compared to all the approaches with the extended data set. 
This is not an issue as SBR benefits were not expected in terms of prediction
accuracy but rather on the enforcing of the monotonicity (see
Sec.~\ref{sec:exp_eval_SBR}).


We are also interested in measuring the results with smaller training sets,
again using MAE as
metric; we keep the test set size fixed at 1000 elements.
Table~\ref{tab:train_size_comparison} reports the experimental results; it has
the 
same structure of Tab.~\ref{tab:mae_comparison}. 
As expected, the prediction accuracy 
decreases with the training set size, but the benefits brought 
by the domain knowledge remain -- over all training set size, the improvement 
brought by the engineered features is 38.7\%.

\begin{table*}
\scriptsize\sf\centering
\begin{tabular}{l cc cc cc cc ccc}
\toprule
\multirow{2}{*}{\emph{Train Set Size}} & \multirow{2}{*}{AutoSklearn} &
\multicolumn{2}{c}{NN-1}  & \multicolumn{2}{c}{NN-2} 
& \multicolumn{2}{c}{NN-3} & \multicolumn{2}{c}{NN-4} & \multirow{2}{*}{GCNN} & 
  \multirow{2}{*}{SBR} \\
 & & Base & Ext. & Base & Ext. & Base & Ext. & Base & Ext. & & \\
\midrule
\emph{500}  & 0.288 & 0.196 & 0.131 & 0.100 & 0.064 & 0.140 & 0.078 & 0.144 &
0.134 & 0.316 & 0.190 \\ 
\emph{1000} & 0.285 & 0.178 & 0.107 & 0.087 & 0.048 & 0.108 & 0.056 & 0.142 &
0.117 & 0.256 & 0.181 \\
\emph{2000} & 0.278 & 0.155 & 0.085 & 0.077 & 0.041 & 0.094 & 0.047 & 0.119 &
0.060 & 0.210 & 0.162 \\
\emph{5000} & 0.274 & 0.126 & 0.076 & 0.057 & 0.033 & 0.081 & 0.037 & 0.086 &
0.067 & 0.166 & 0.133 \\
\bottomrule
\end{tabular}
\caption{Knowledge injection approaches comparison: average on all benchmarks
MAE -- varying training set size}
\label{tab:train_size_comparison}   
\end{table*} 

\subsection{Semantic Based Regularization Impact}
\label{sec:exp_eval_SBR}
This section provides additional details on the experiments on
data augmentation and SBR. 
The model was tested on the previous benchmarks and different ratios of data
injected, 
i.e. 25\% and 75\%.
In order to have a more precise evaluation of the approach, we relied on another
metric beside MAE,
that is the number of violated monotonicity constraints -- the goal of this
approach is to \emph{reduce} their 
number.
We underline that not every benchmark had monotonicity issues (as they are
outliers),  and
in these cases the 
regularization factor is of no use and might keep the model from a good
approximation. 
For this reason, Table~\ref{tab:sbr_MAE_comparison} and
Table~\ref{tab:sbr_Viol_comparison}, report just
the values from significant benchmarks 
(i.e. benchmarks that exhibit the most marked non-monotonic behavior), these are
\emph{convolution} and \emph{correlation}. The third column reports the result
obtained with NN-1 without the additional features. Columns 4-6 correspond to
the 
results obtained with data augmentation and SBR, with different percentages of 
injected data (0\%, 25\%, 75\%)

\begin{table}[H]
\scriptsize\sf\centering
\begin{tabular}{l cc cc cc cc cc}
\toprule
\emph{Benchmark} & \emph{Size} &\multicolumn{1}{c}{\emph{NN-1}} &
\multicolumn{1}{c}{SBR}  & \multicolumn{1}{c}{\emph{SBR 25\%}} &
\multicolumn{1}{c}{\emph{SBR 75\%}} \\
 &  & MAE  & MAE & MAE & MAE \\
\midrule
\multirow{2}{*}{\emph{convolution}} & $500$  & 0.012 & 0.019 & 0.013 & 0.011 \\
                                    & $5000$ & 0.005 & 0.006 & 0.006 & 0.006 \\ 
\midrule
\multirow{2}{*}{\emph{correlation}} & $500$  & 0.263 & 0.265 & 0.263 & 0.262 \\ 
                                    & $5000$ & 0.139 & 0.059 & 0.059 & 0.139 \\
\bottomrule
\end{tabular}
\caption{SBR: MAE -- comparison of small and large training sets}
\label{tab:sbr_MAE_comparison}  
\end{table} 

\begin{table}[H]
\scriptsize\sf\centering
\begin{tabular}{l cc cc cc cc cc}
\toprule
\emph{Benchmark} & \emph{Size} &\multicolumn{1}{c}{\emph{NN-1}} &
\multicolumn{1}{c}{SBR}  & \multicolumn{1}{c}{\emph{SBR 25\%}} &
\multicolumn{1}{c}{\emph{SBR 75\%}} \\
 &  & \#Viol.  & \#Viol. & \#Viol. & \#Viol. \\
\midrule
\multirow{2}{*}{\emph{convolution}} & $500$  & 168  & 156   & 171  & 126 \\
                                    & $5000$ & 0    & 13    & 12   & 6  \\ 
\midrule
\multirow{2}{*}{\emph{correlation}} & $500$  & 111  & 120   & 116  & 98  \\ 
                                    & $5000$ & 91   & 59    & 71   & 92 \\ 
\bottomrule
\end{tabular}
\caption{SBR: number of violated constraints (\#Viol.)}
\label{tab:sbr_Viol_comparison} 
\end{table} 

With larger training sets, the benefits of data augmentation and SBR are
marginal: 
the additional constraint on the loss function is not very useful, given the
abundance of 
training samples allowing for better generalization. Similarly, larger training
sets lead
to a natural decrease in the number of monotonicity constraints violated (as
their 
proportion in the training set diminishes).
Nevertheless, the more interesting results can be observed when fewer data
points are 
available, since the models show a decrease in 
the number of violated constraints opposed to 
the network without regularization. Furthermore, the
networks performed better with higher ratios of data injected, i.e. 18\%, on
average.
Finally, the MAE seems to have values compatible to
the results obtained with \emph{NN-1}, a good result since prediction accuracy
was not SBR's scope.
These results encourage the 
idea of a hybrid model merging data augmentation plus SBR and additional
features 
(both approaches enabled by the injection of domain knowledge),
as future development of this work.

\section{Conclusion \& Future Works}
\label{sec:conclusion}
In this paper we present a strategy for injecting domain knowledge in a 
ML model. As a case of study, we considered a learning task from the
transprecision computing field,
namely predicting the computation error associated to the precision used for 
handling a set of FP variables composing a benchmark. This is a difficult
regression problem,
hard to be addressed with pure data-driven ML methods; we have shown how
critical 
improvements can be reached by injecting domain knowledge in the ML models.

We introduced three knowledge-injection approaches and applied them on top
of NNs with varying structures: feature engineering, a GCNN, and a data
augmentation
scheme enabled by SBR.
The GCNN approach did not improve the accuracy of the ML model w.r.t. the
baseline 
and it should be explored more in detail. Conversely, the creation of extended 
data set was revealed to be extremely useful, leading to remarkable reduction in
prediction error (39.7\% on average and up to 47.5\% in the best case). 
Data augmentation plus SBR showed its potential with training sets of 
limited size, in terms of reduced number of violated monotonicity constraints
while 
preserving the ML models' prediction accuracy.

In future works we plan to integrate the learners in an optimization model for
solving the FP tuning precision problem. In this regard we will explore active
learning
strategies and we expect SBR to have good result, especially when combined with 
the additional features (the methods are orthogonal). Moreover, we will perform 
experiments with other domain knowledge injection approaches, for instance by
building data sets in accordance with the prior information and by exploiting
the 
knowledge to guide the training of the NN by constraining its output.
 
\section*{Acknowledgments}
This work has been partially
supported by European H2020 FET project OPRECOMP (g.a. 732631).

\FloatBarrier

\bibliographystyle{alpha}
\bibliography{bib}

\end{document}